\documentclass[sigconf]{acmart}

\usepackage{booktabs} % For formal tables

\acmArticle{4}

\begin{document}

\copyrightyear{2018}
\acmYear{2018}
\setcopyright{acmlicensed}
\acmConference[SAC 2018]{SAC 2018: Symposium on Applied Computing }{April 9--13, 2018}{Pau, France}
\acmPrice{15.00}
\acmDOI{10.1145/3167132.3167165}
\acmISBN{978-1-4503-5191-1/18/04}

\title{Deep Reinforcement Learning Boosted by External Knowledge}

\author{Nicolas Bougie}
\affiliation{%
  \institution{National Institute of Informatics}
  \streetaddress{2-1-2 Hitotsubashi, Chiyoda}
  \city{Tokyo} 
  \country{Japan} 
  \postcode{101-8430}
}
\email{nicolas-bougie@nii.ac.jp}

\author{Ryutaro Ichise}
\affiliation{%
  \institution{National Institute of Informatics}
  \streetaddress{2-1-2 Hitotsubashi, Chiyoda}
  \city{Tokyo} 
  \country{Japan} 
  \postcode{101-8430}
}
\email{ichise@nii.ac.jp}
% The default list of authors is too long for headers}
\renewcommand{\shortauthors}{N. Bougie et al.}

\begin{abstract}
Recent improvements in deep reinforcement learning have allowed to solve problems in many 2D domains such as Atari games.
However, in complex 3D environments, numerous learning episodes are required which may be too time consuming or even impossible especially in real-world scenarios. We present a new architecture to combine external knowledge and deep reinforcement learning using only visual input. A key concept of our system is augmenting image input by adding environment feature information and combining 
two sources of decision.
We evaluate the performances of our method in a 3D partially-observable environment from the Microsoft Malmo platform. Experimental evaluation exhibits higher performance and faster learning compared to a single reinforcement learning model.
\end{abstract}




\maketitle

\section{Introduction}
Reinforcement learning is a technique which automatically learns a
strategy to solve a task by interacting with the environment and learning
from its mistakes. 
By combining reinforcement learning and deep learning to extract features from the input, a wide variety of tasks such as Atari 2600
games \cite{2013arXiv1312.5602M} are efficiently solved. 
However, these techniques applied to 2D domains struggle in complex environments such as three-dimensional virtual worlds resulting a prohibitive training time and an inefficient learned policy.

\noindent A powerful recent idea to tackle the problem of computational expenses is to modularise the models into an ensemble of experts \cite{lample2017playing}. Since each expert focuses on learning a stage of the task, the reduction of the actions to consider leads to a shorter learning period. Although this approach is conceptually simple, it does not handle very complex environments and environments with a large set of actions.

A similar idea of extending the information extracted from low-level architectural modules \cite{lieto2017knowledge} with high-level ones have been previously used in the area of cognitive systems \cite{lebiere2017} but does not directly relies on RL and was limited to a supervised classification problem. The idea was to leverage information about videos with external ontologies to detect events in videos.

Another technique is called \textit{Hierarchical
Learning} \cite{tessler2017deep}\cite{Barto03recentadvances} and is used
to solve complex tasks, such as "simulating human brain" \cite{lake2016building}. It is inspired by
human learning which uses previous experiences to face new
situations. Instead of learning directly the entire task, different sub-tasks
are learned by the agent. By reusing knowledge acquired from the previous
sub-tasks, the learning is faster and easier. Some limitations are the necessity to re-train the model which is time consuming and problems related to catastrophic forgetting of knowledge on previous tasks.

In this paper, our approach focuses on combining deep reinforcement learning
and external knowledge.
Using external knowledge is a way to supervise
the learning and enhance information given to the agent by introducing human
expertise. We augment the input of a reinforcement learning model whose input is raw pixels by adding high-level information created from simple knowledge about the task and recognized objects. We combine this model with a knowledge based decision algorithm using Q-learning \cite{Watkins1992}.
In our experiments, we demonstrate that our framework successfully learns in
real time to solve a food gathering task in a 3D partially observable environment by only using visual inputs. We evaluate our technique on the Malmo platform built on top of a 3D virtual environment, Minecraft. Our model is especially suitable for tasks involving navigation, orientation or exploration, in which we can easily provide external knowledge.

The paper is organized as follows. Section 2 gives an
overview of reinforcement learning and most recent models. The environment is presented in Section 3. The main
contribution of the paper is described in Sections 4. Results are presented
in Section 5. Section 6 presents the main conclusions drawn from the work.

\section{Related Work}
Below we give a brief introduction to reinforcement learning and the models used into our system architecture. 

\subsection{Reinforcement Learning}
Reinforcement learning consists of an agent learning a policy by interacting with an environment. At each time-step the agent receives an observation $s_{t}$ and choose an action $a_{t}$. The agent gets a feedback from the environment called a reward $r_{t}$. Given this reward and the observation, the agent can update its policy to improve the future rewards.
\\Given a discount factor $\gamma$, the future discounted reward, called return $R_{t}$, is defined as follows : 

\begin{equation}\label{eq:eq}
R_{t}=\sum_{t'=t}^{T} \gamma^{t'-t}r_{t'}
\end{equation}

\noindent The goal of reinforcement learning is to learn to select the action with the maximum return $R_{t}$ achievable for a given observation \cite{sutton1998reinforcement}.
From Equation (\ref{eq:eq}), we can define the action value $Q^{\pi}$ at a
time $t$ as the expected reward for selecting an action $a$ for a given
state $s_{t}$ and following a policy $\pi$.

\begin{equation}
Q^{\pi}(s,a) = \mathbb{E} \left\lbrack R_{t}\mid s_{t}=s,a \right\rbrack
\end{equation}

\noindent The optimal policy is defined as selecting the action with the optimal Q-value, the highest expected return, followed by an optimal sequence of actions. This obeys the Bellman optimality equation: 

\begin{equation}\label{eq:opti-bellman}
Q^{*}(s,a) = \mathbb{E} \left\lbrack r+ \gamma \max_{a'}Q^{*}(s^{'},a^{'})\mid s,a \right\rbrack 
\end{equation}

\noindent When the state space or the action space is too large to be represented, it is possible to use an approximator to estimate the action-value $Q^{*}(s,a)$: 

\begin{equation}
Q^{*}(s,a)\approx Q(s,a;\theta)
\end{equation}

\noindent Neural networks are a common way to approximate the action-value. The parameters of the neural network $\theta$ can be optimized to minimize a loss function $L_{i}$ defined as the expected temporal difference error of Equation (\ref{eq:opti-bellman}): 

\begin{equation}
L_{i}(\theta_{i})= \mathbb{E}_{s,a,r,s^{'}}  \left\lbrack (y_{i}-Q(s,a;\theta_{i}))^{2} \right\rbrack 
\end{equation}

\noindent where
$y_{t} = r_{t} +\gamma \max_{a^{'}} Q_{\theta_{target}}(s_{t+1},a^{'})  $

\noindent The gradient of the loss function with respect to the weights is the following :  

\begin{equation}
\scalebox{0.8}{%
$\nabla_{\theta{i}}L_{i}(\theta{i})=\mathbb{E}_{s,a,r,s^{'}}  \left\lbrack (r+\gamma\max_{a^{'}}Q(s^{'}, a^{'};\theta_{i-1}) - Q(s, a;\theta_{i})) \nabla_{\theta{i}}Q(s, a;\theta_{i}) \right\rbrack$}
\end{equation}

\noindent Mnih \textit{et al.} (2013) used this idea and created the famous method called Deep
Q-learning (DQN) \cite{2013arXiv1312.5602M}. However, the learning may be slow due
to the propagation of the reward to the previous states and actions.

\noindent Similarly, the value function $V^{\pi}(s)$ which represents the expected return for a
state $s$ following a policy $\pi$ is defined as follows:

\begin{equation}
V^{\pi}(s)=  \mathbb{E}  \left\lbrack R_{t}\mid s_{t} = s \right\rbrack 
\end{equation}

\noindent Some reinforcement learning models such as \textit{Actor-Critic} or \textit{Dueling Network} decompose the Q-values $Q(s,a)$ into two more fundamental values, the value function $V(s)$ and the advantage function  $A(a,s)$ which is the benefit of taking an action compared to the others.

\begin{equation}
A(s,a)=Q(s,a)-V(s)
\end{equation}
\subsection{Asynchronous Advantage Actor-Critic (A3C)}
It was shown that combining methods of deep learning and reinforcement learning is very unstable.
To deal with this
challenge, many solutions store the agent's data into a memory, then the
data can be batched from the memory. It is done because sequences of data
are highly correlated and can lead to learn from its mistakes resulting in a
worse and worse policy.
A3C \cite{mnih2016asynchronous} avoids computational and memory problems by
using asynchronous learning. It allows the usage of on-policy reinforcement learning algorithms
such as \textit{Q-learning} \cite{Watkins1992} or \textit{advantage
actor-critic}. The learning is stabilized without using experience replay
and the training time is reduced linearly in the number of learners.

\begin{figure}[tb]
\centering
\includegraphics[width=0.9\linewidth]{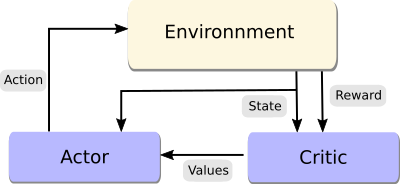}
\caption{Actor-critic model}
\label{fig:actor-critic-illustration}
\end{figure}

The learners of A3C which use their own copy of the environment are trained in parallel. Each
process will learn a different policy and hence will explore the environment
in a different way leading to a much more efficient exploration of the
environment than with a replay memory.
A process updates its own policy based on an advantage actor-critic model \cite{Konda1999ActorCriticA} (Figure
\ref{fig:actor-critic-illustration}). The actor-critic model is composed by an actor
which acts out a policy and a critic which evaluates the policy. The main thread is updated
periodically using the accumulated gradients of the different processes.

\noindent The critic takes as input the state and the reward and outputs a score
to criticize the current policy. In the case of advantage actor-critic
model, the critic estimates the advantage function which requires to
estimate $V$ and $Q$.

The actor does not have access to the reward but only to the state and the
advantage value outputted by the critic. Contrary to the critic which is
value based, the actor directly works into the policy space and changes the policy
towards the best direction estimated by the critic.
Optimization techniques such as stochastic gradient descent are used to find
$\theta$ that maximizes the policy objective function $J(\theta)$.
The policy gradient objective function $\nabla_{\theta}J(\theta)$ is defined as follows: 

\begin{equation}
\nabla_{\theta}J(\theta) =  \mathbb{E}_{\pi,\theta}  \left\lbrack \nabla_{\theta}log\pi_{\theta}(s,a)A^{w}(s,a)\right\rbrack
\end{equation}

\noindent where $A^{w}(s,a)$ is a long term estimation of the reward to allow the actor to go in the direction that the critic considers the best.

\subsection{Dueling Network}

\begin{figure}[tb]
\centering
\includegraphics[width=50mm, bb= 0 0 250 165]{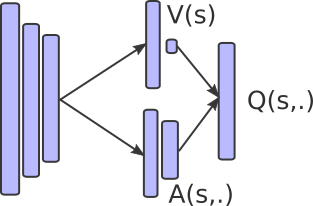}
\caption{Dueling network architecture}
\label{fig:dueling-network-illustration}
\end{figure}

The idea is to separately compute the advantage function and the value
function and combine these two values at the final layer (Figure
\ref{fig:dueling-network-illustration}). The dueling network \cite{2015arXiv151106581W} may not need to care about both values and
the advantage at any given time.
The estimation of a state value is more robust by decoupling it from the
necessity of being attached to a specific action. This is particularly
useful in states where its actions do not affect the environment in any
relevant way. For example, moving left or right only matters when a
collision is upcoming.
The second stream, which estimates the advantage function values, is relevant when the model needs to make a choice over the actions in a state.
The Bellman's equation (\ref{eq:opti-bellman}) becomes now:

\begin{equation}
Q(s,a;\theta,\alpha,\beta) = V(s,\theta,\beta) + (A(s,a;\theta,\alpha) - \max_{a^{'}\in \mathcal{A}}A(s,a^{'};\theta,\alpha))
\end{equation}

\noindent And by changing the max by a mean: 

\begin{equation}
Q(s,a;\theta,\alpha,\beta) = V(s,\theta,\beta) + (A(s,a;\theta,\alpha) - \frac{1}{|\mathcal{A}|}\sum_{a^{'}} A(s,a^{'};\theta,\alpha))
\end{equation}

\noindent With $\theta$ the shared parameters of the neural network, $\alpha$ the parameters of the stream of the advantage function $\mathcal{A}$ and $\beta$ the parameters of the stream of the value function V. Since the output of the two streams produces a Q function, it can be trained
with many existing algorithms such as Double Deep Q-learning
(DDQN) \cite{van2016deep} or SARSA \cite{rummery1994line}. The main
advantage is that for each update of the Q-values, the value function is
updated whereas with traditional Q-learning only one action-value is
updated.

\section{Task \& Environment}
\begin{figure}[tb]
\centering
\includegraphics[width=.7\linewidth, bb= 0 0 400 400]{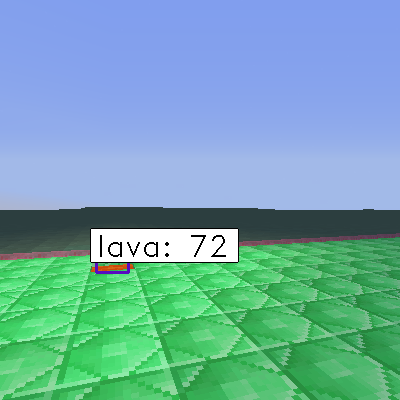}
\caption{Screenshot of the environment}
\label{fig:object-reco-yolo}
\end{figure}

We built an environment on the top of the Malmo platform \cite{malmoPlatform} to evaluate our idea. Malmo is an open-source platform that allows us to create scenarios with Minecraft engine. To test our model, we trained an agent to collect foods in a field with obstacles. The agent can only receive partial information of the environment from his viewpoint. We only use image frames to solve the scenario. An example of screenshot with the object recognition results is shown in Figure \ref{fig:object-reco-yolo}.

The goal of the agent is to learn to have a healthy diet. It involves to recognize the objects and learn to
navigate into a 3D environment. The task consists in picking up food from
the ground for 30 seconds. Food is randomly spread across the environment and four obstacles are randomly generated. Each of the 20 kinds of food has an associated reward when the agent picks
it up. This reward is a number between +2 (healthy) and -2 (unhealthy). They
are distributed equitably, meaning that a random agent should get a reward
of 0. 

The settings were: window size: \textit{400 $\times$ 400} pixels, actions: \textit{turn
left, turn right, crouch, jump, move straight and move back} , number of
objects: \textit{200},  number of obstacles: \textit{4}.
The actions \textit{turn left} and \textit{turn right} are continuous actions to make the learning smoother as consecutive frames are more similar.

\section{System Architecture}
\subsection{General Idea}
\begin{figure}[tb]
\centering
\includegraphics[width=1.0\linewidth]{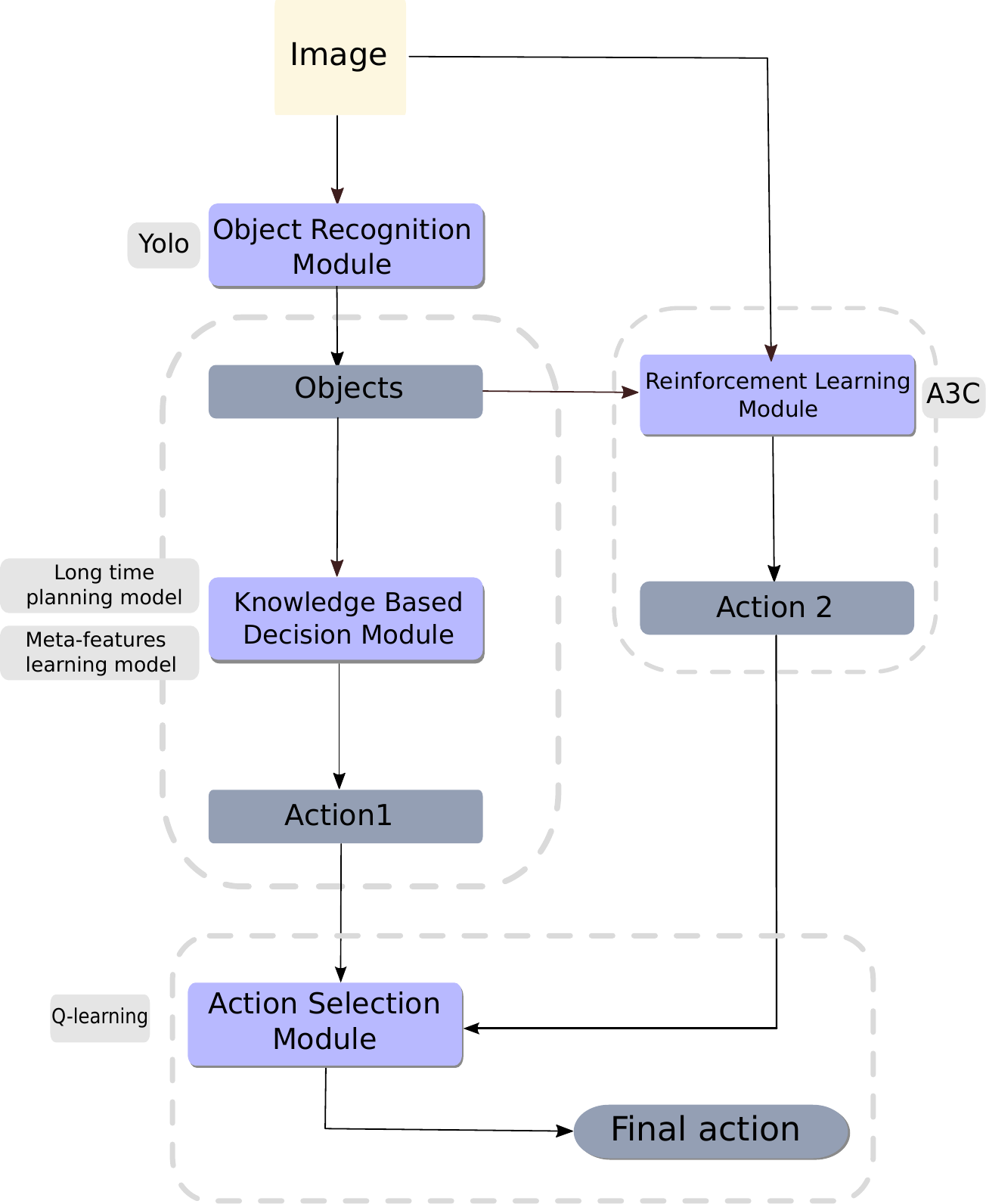}
\caption{Global architecture of DRL-EK}
\label{fig:new_framework_image}
\end{figure}

Figure \ref{fig:new_framework_image} describes the global architecture of our
new framework called DRL-EK. It consists of four modules: an Object Recognition
Module, a Reinforcement Learning Module, a Knowledge Based Decision Module, and
an Action Selection Module.

The object recognition module identifies the objects within the current image and generates high-level features.
These features of the environment are then used to augment the raw image input to the reinforcement learning module.
In parallel, the knowledge based decision module selects another action by combining external knowledge and the object recognition module outputs.
To manage the trade-off between these two sources of decision we use an action selection module. The chosen action is then acted by the agent and the modules are updated from the obtained reward.

\subsection{Object Recognition Module}
Injecting external knowledge requires to understand the scene at a high-level in order to be interpreted by a human. The easiest way to understand an
image is to identify the objects. For example, it is intuitive to give more
importance to the actions $turn$ or $jump$ than the action $move\ straight$
when an obstacle is in front of the agent.
To recognize the objects, the module uses You Only Look Once (YOLO) \cite{redmon2016you}\cite{2016arXiv161208242R} library which is based on a deep convolutional neural network.
As input, we use an RGB image of size 400$\times$400 pixels. YOLO predicts in real time the bounding boxes, the labels and
confidence scores between 0 and 100 of the objects. An example is shown in Figure \ref{fig:object-reco-yolo}.
We trained YOLO on a dataset of 25 000 images with twenty different classes corresponding to the food that is presented in the environment.

The model is trained off-line before starting the learning into the environment. The neural network architecture is adapted from the one
proposed by Redmon \textit{et al.} (2016) for the Pascal VOC dataset \cite{2016arXiv161208242R}.
In order to recognize small objects, the size of cells is decreased from 7 to 5 pixels and the number of bounding boxes for each cell is increased from 2 to 4.

In addition to the identified objects, the module creates feature information about the current frame. To generate these high-level abstraction features we combine the recognized objects and external knowledge. They are then used as input by the reinforcement learning module and the knowledge based decision module. We designed two types of features \textit{presence of objects} and \textit{important area}.

\subsubsection{Presence Of Objects Features}

The first type of features is a vector of booleans which 
indicates whether an object appears or not within the current image.
The size of this vector is the number of different objects in the environment. 
Since some objects are not helpful to solve the task, we can decide to only take some of the objects into account based on our knowledge about the task.

\begin{figure}[tb]
\centering
\includegraphics[width=0.6\linewidth]{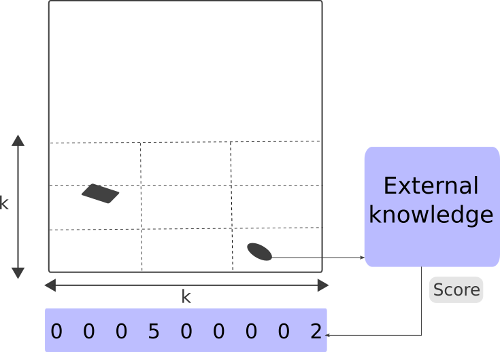}
\caption{Important areas of an image}
\label{fig:Importantareas}
\end{figure}
\begin{figure}[tb]
\centering
\includegraphics[width=1.0\linewidth]{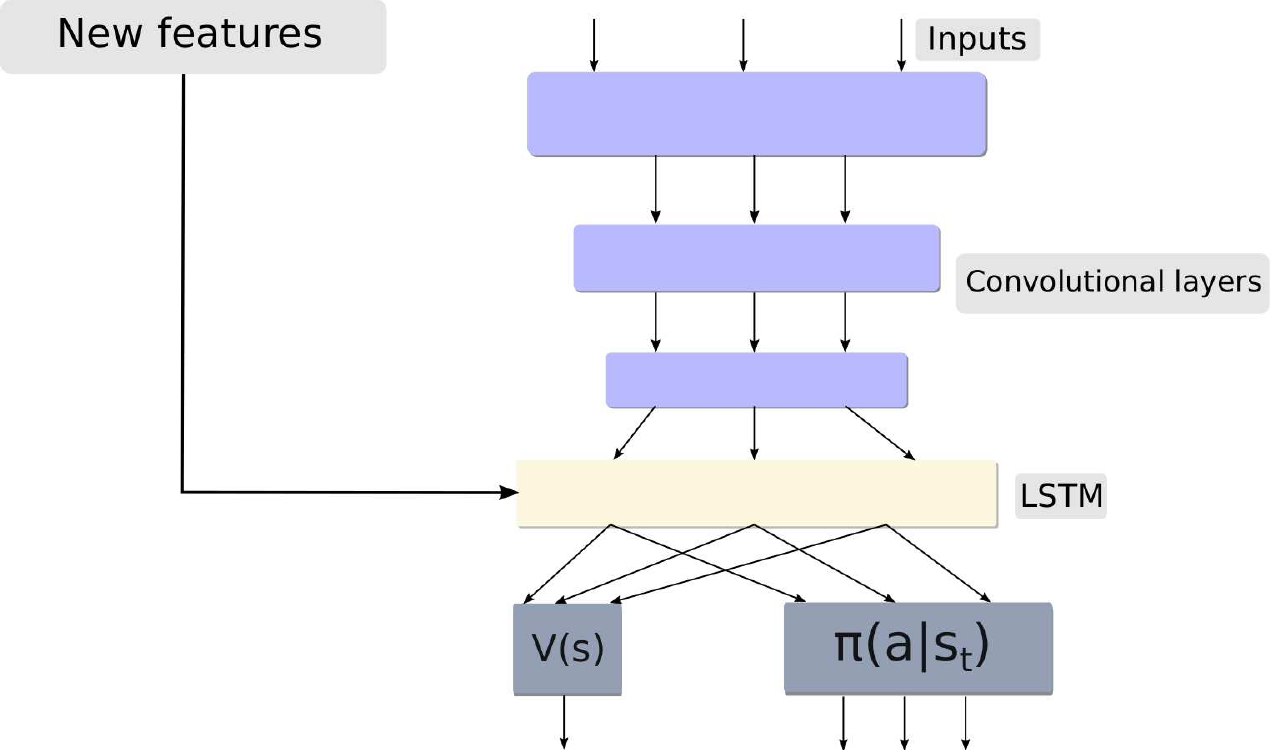}
\caption{Injection of new features into the reinforcement learning module (A3C)}
\label{fig:Basic_external_knowledges_injection}
\end{figure}

\subsubsection{Important Area Features}

As the position of objects is important, we encode information about objects within each area of the image. We split the image into $k$ rectangles vertically and horizontally. So, the number of areas is $k^{2}$ and for each one we compute a score (Figure \ref{fig:Importantareas}).
The score of an area is the sum of the score of the objects within this
area. 
External knowledge can be introduced by shaping the score of the objects. From our knowledge about the task, we manually defined the scores to indicate whether or not an object is important to solve the task.
To tackle problems with partially observable environments, we keep track of recent information by concatenating the array of scores of the current
frame with the arrays of the two previous frames. 

In our experiments, the top half of the images only contains the sky so we computed the important area features on the
half bottom of the images. We gave a score of -15/+5 to foods we think is unhealthy (cake,cookie) / healthy (meat, fruit) and 0 for the others. That way, if an area contains a healthy food such as a fruit and a sweet food,
then the score of the area will be lower than an area containing only a fruit or no object.
We set the number of rectangles to 3 (9 areas in total: 3$\times$3). We found that
with a higher number of areas the amount of encoded information is bigger but
information quality of each area is worse than with 3 areas.

\subsection{Reinforcement Learning Module}

For a computer, learning from an image is difficult and requires a lot of training steps. To deal with it, the entry point of most of the reinforcement learning models is a recurrent convolutional neural network \cite{hausknecht2015deep} to extract temporal and spatial features of the image.

We trained a deep reinforcement learning model to perform policy learning and we modified the neural network structure to incorporate external knowledge. In addition to the image input, we injected \textit{
presence of objects} or \textit{important area} features which are created by the object recognition module. In the neural network, we give to a Long Short Term Memory (LSTM) \cite{Hochreiter} the output of the last convolutional layer concatenated with the new features (Figure
\ref{fig:Basic_external_knowledges_injection}). The next layers of the neural network are two separated fully-connected layers to estimate the value function $V(s)$ and the policy $\pi(a|s_{t})$.
The purpose is to help the model at the beginning of the training to recognize and focus on objects. The new features augment the raw image input to the reinforcement learning model by adding high-level information. For example, from presence of objects features the model can decide which actions are allowed or not. If a $door$ is detected some of the
actions may become irrelevant such as $jumping$.

The choice of the reinforcement learning model highly depends on the
environment. Since the model at each time-step takes an input and outputs an action, we can easily substitute most of the reinforcement learning techniques such as Deep Q-learning (DQN) \cite{2013arXiv1312.5602M}, Deep Deterministic Gradient Policy (DDPG) \cite{2015arXiv150902971L}, Dueling Network \cite{2015arXiv151106581W} or Asynchronous Actor-Critic Agents(A3C) \cite{mnih2016asynchronous} by using a recurrent convolutional neural network as
state approximator.

A3C is the most suitable model to solve our task. We tested and empirically searched the best parameters such as a good convolutional neural network architecture and the choice of the optimizer of this model.
It provides a baseline to evaluate the importance of each module of our architecture on the final policy.

Working directly with $400\times400$ pixel images is too computationally
demanding. We apply image preprocessing before training A3C. The raw frame is resized to $200\times200$ pixels. To
decrease the storage cost of the images we convert the image scale from
$0-255$ to $0-1$.

We set the number of workers of A3C to 3 and
a convolutional recurrent neural network is used to approximate the states. The reason why we use a recurrent neural network is because the environment is partially observable.
The input of the neural network of A3C estimator consists in a 200$\times$200$\times$3
image. The 4 first layers convolve with the following parameters (filter:
32,32,32,32, kernel size: 8$\times$8,4$\times$4,3$\times$3,2$\times$2, stride size: 2,2,2,1) and apply
a rectifier nonlinearity.
It is followed by a LSTM layer of size 128 to incorporate the temporal features of the environment.
Two separate fully connected layers predict the value function and a policy function, a distribution of probability over the actions.
We use RMSProp \cite{hinton2012neural} as optimization technique with $\epsilon = 10^{-6}$ and minibatches of size 32 for training.

\subsection{Knowledge Based Decision Module}
%dire que prend pas en compte image
We believe that the agent is not able to accurately understand and take into
account the objects of the environment. A human can easily understand and
make a decision from high-level features such as the utility or name of an
object. Getting this level of abstraction is difficult but we can
help the machine by giving it less low-level information such as color of
pixels but more high-level information such as the importance of an area of
the image.

Moreover, when the reinforcement learning module is fed with the images and the presence of objects or important areas features, the
training time is long due to the size of state space.
The knowledge based decision module is able to select an action using external knowledge and high-level features generated by the object recognition module and without direct access to the image.
We propose two different approaches, a long time planning model or a meta-feature learning model.

\subsubsection{Long Time Planning Model}

Our approach to solving the task is based on planning a sequence of actions. We
designed and developed a long time planning model without learning which
combines traditional test-case algorithms and planning.
The model takes as input the probabilities and the bounding
boxes of the objects detected within the current frame.

We store in an array the sequence of planned actions. At each time-step, the model checks if the previously planned sequence of actions
is still the optimal one and if it is not the case (for example the next action
is \textit{jump} but there is no obstacle) the algorithm updates it, otherwise the first action in the array is returned.

To update or plan a sequence of actions, the model uses the information about the objects and manually created rules. First, a test-case verification selects the possible actions. 
An example of simple rule is, if the object \textit{cookie} is on the left of the image then the action \textit{turn left} is forbidden.
Then, to decide of
the action among the remaining actions we use a priority list. If the
selected action is related to the movement, the model estimates the best angle
and the necessary number of steps to perform it.
Finally, the first of the planned actions is returned.

To decrease the number of rules we discretized the image space into four areas: \textit{center, left, right, other}.
We designed 43 rules to prevent the agent from going in the direction of the food we think is dangerous. 
To avoid static behaviour, we give more priority to the actions \textit{turn
left, turn right, move straight} than the others in the priority list.

\subsubsection{Meta-feature Learning Model}
%a refaire Instead of manually creating the rules, we would like to automatically select the optimal action
In our previous approach, we manually create rules to reason on high-level features. To automatically learn the rules and select the optimal action from them, we use a deep reinforcement learning model such as DQN or dueling network. Unlike the reinforcement learning module which uses the image, the only input is high-level features such as \textit{important areas} or \textit{presence of objects}. 
As the input is much smaller than an image, a simple neural network can be trained to approximate the states. The smaller number of parameters leads to a 
faster learning than a model trained from visual information. 

In experiments, we trained a dueling network combined with a double deep Q-learning (DDQN). It empirically
gives a smoother learning than most of the other reinforcement learning
models.
A neural network approximates the states. It consists in 3 fully connected layers of size 100 with a rectifier nonlinearity activation function. Network was trained using the $Adam$ algorithm \cite{journals/corr/KingmaB14}, learning rate of $10^{-3}$ and minibatches of size 32.
As input, we used a slightly modified version of the important area features outputted by the object recognition module. To create important area features, we filtered the objects too far and the objects with a
confidence score less than 0.25. Taking into account an object
such as \textit{grass} is irrelevant and makes the learning more difficult.
We only used dangerous or very healthy (10 objects out of
20) objects and removed 2 objects that we know are difficult to
distinguish.

\subsection{Action Selection Module}

The module aggregates the actions proposed by the reinforcement
learning module and the knowledge based decision module to select the action that the agent will perform in the environment.
The goal is to
take advantage of the fast learning of the knowledge based decision module
and the quality of the policy learned by the reinforcement learning module. 
An important aspect of the action selection, is selecting an action with the highest expected return but also detecting error patterns to correct them. An error must be detected when an action which has not been proposed could offer a higher \textit{
return}.

This is achieved by training a Deep Q-learning model to select the best action, detect and
correct the error patterns. There is no restriction on the possible actions meaning
that the final action may be different from the two proposed actions if an error is detected. 
We encode the proposed action by the two modules into two indicator vectors. An indicator vector is a binary vector with only one unit turned on to indicate the recommended action. The neural network input is the concatenation of these two vectors.

%dire ce qu'est tau et que proba de selectionner action dans un etat s
The Q-learning algorithm is trained using a
Boltzmann distribution (Equation \ref{eq:boltzmann}) as explorer and experience replay (each experience is stored into
memory and the algorithm is run on randomly sampled batch) with a memory of
size $10^{6}$. Equation \ref{eq:boltzmann} gives the probability of selecting an action in a given state $s$.
\begin{equation}\label{eq:boltzmann}
P_{s}(a)=\frac{exp(Q(s,a)/\tau)}{\sum_{a^{'}\in A}exp(Q(s,a^{'})/\tau)}
\end{equation}
The Q-network is composed of 2 hidden fully connected layers of size 50 and are followed by rectified linear units.

\section{Experiments}
We conducted several experiments for evaluating our architecture. In all our experiments, we set the discount factor to 1.0. According to our different tests, on
average the best reward that a perfect agent can get in 30 seconds is 9.

\subsection{Object Recognition}
\begin{figure}[tb]
\centering
\includegraphics[width=1.0\linewidth, bb= 0 0 493 343]{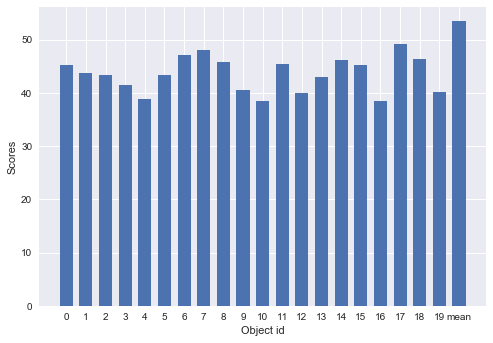}
\caption{Average precision over all the classes obtained by the object recognition module \label{map-scores}}
\label{fig:yolo-output}
\end{figure}

We evaluated our object recognition module for
understanding the correctness of obtained object information in the
environment.
Figure \ref{fig:yolo-output} reports the object recognition module
performance.
%Figure \ref{fig:object-reco-yolo} shows an example of the output of the system.
In this experiment, we measured the mean average precision (mAP) as the error metric. The
results are similar to the results presented by the authors (Redmon \textit{et al.},
2017) \cite{2016arXiv161208242R} on the Pascal
VOC. dataset \cite{Everingham:2010:PVO:1747084.1747104}. We obtained a mean average precision of $53.47$. Although other libraries could offer higher performance, the real-time detection was the main criterion for selecting YOLO.

We noticed that most of the errors are false positives $(68.3\%)$ whereas the false
negatives $(31.7\%)$ are uncommon. It leads to an agent with a policy more greedy and safer. 
As shown in the figure, the average precision is similar for every class. The performance of YOLO is slightly affected by the complexity of the
objects such as their shape, color or size.

\subsection{Long Time Planning Model}

\begin{figure}[tb]
\centering
\includegraphics[width=1.0\linewidth, bb= 0 0 493 344]{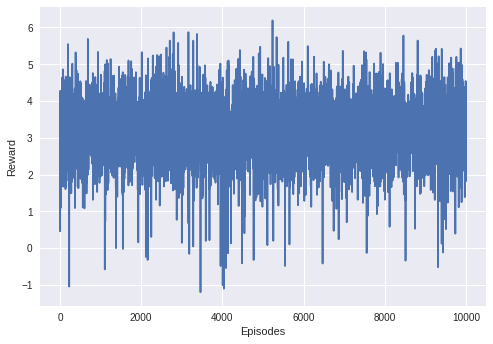}
  \caption{Evolution of the reward of the long time planning model}
  \label{fig:planning-decision-res}
\end{figure}
Next, we tested the long time planning model in the knowledge based decision module to evaluate the effectiveness of this approach. 
We
only utilized the object recognition module and the knowledge based decision module in
our framework. The long time planning model
performs much better than a random agent with an average
reward of 3.3 (Figure \ref{fig:planning-decision-res}), since expected reward of random agent is 0.

The most likely cause of the wide variance is the difficulty to handle all possible cases with manually created rules. For instance, the agent has difficulty in gathering food near obstacles.
Since there is no learning, the quality of the agent only depends of the quality of the rules and is not able to converge. On the other hand, from the first episode the average reward is much higher than any other learning based models.

\subsection{Meta-feature Learning Model}
\begin{figure}[tb]
\includegraphics[width=1.0\linewidth, bb= 0 0 1394 354]{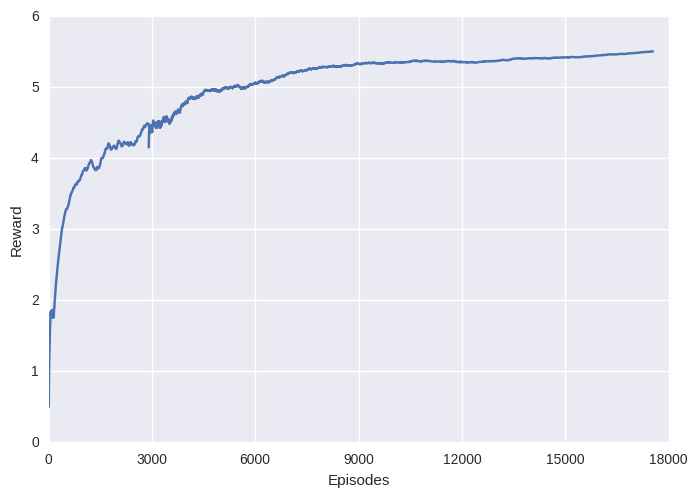}
\caption{Average rewards of the meta-feature learning model with different parameters. The rewards were averaged over 200 episodes after 5000 training episodes}
\label{fig:meat-feat-optimization}
\end{figure}

To evaluate performance of the meta-feature learning model we use as baseline the long time planning model. We optimized its parameters,
by sampling hyper-parameters from categorical distributions:

\begin{itemize}
\item Number of areas sampled from $\{4,9,16,25\}$ 
\item Number of hidden layers from $[1,5]$
\item Size of hidden layers sampled from $\{25,50,100,200,300\}$ 
\end{itemize}
Figure \ref{fig:meat-feat-optimization} reports an example of hyper-parameter optimization results. Each cell corresponds to a configuration of parameters.
As can be seen on the figure, a number of hidden layers larger than two or a large number of areas results in lower performance.
The best hyper-parameters are $9$ areas, and a neural network with 3 hidden
fully-connected layers of size 100. 
Training time is about 4 hours for each configuration on a Nvidia Titan-X GPU.
%As can be seen in

Figure \ref{fig:figure7} shows how the average total reward of
the meta-feature learning model evolves during training with the optimal
settings.
The dueling network architecture effectively learns to solve the task from the important area features. The learning is fast during the first 3000 episodes and the
average reward quickly converges around 5.4.
It is also interesting to note that this approach rapidly achieves higher performance than the long time planning model. Automatic rule learning is more effective than manual rule construction. Unfortunately, the rules cannot be represented in a human-interpretable way. 

\begin{figure}[tb]
\centering
\includegraphics[width=1.0\linewidth, bb= 0 0 502 358]{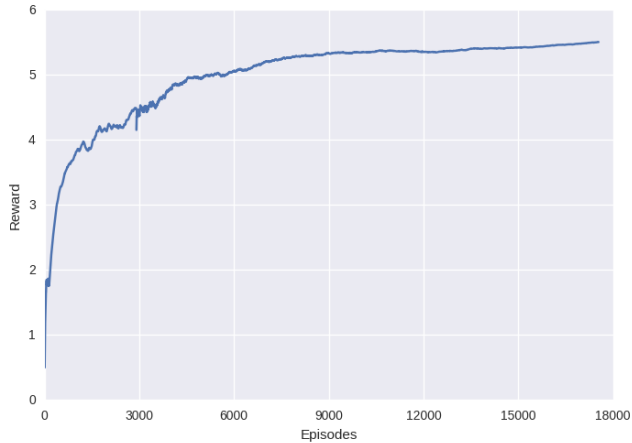}
\caption{Average reward using meta-feature learning}
\label{fig:meta-feat_learning}
\end{figure}

\subsection{Action Selection}
\begin{figure}[tb]
\centering
\includegraphics[width=1.0\linewidth, bb= 0 0 513 358]{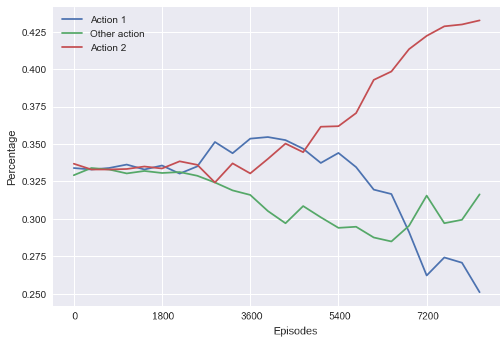}
\caption{Frequency of selection of each action
\label{fig:actin selection}}
\end{figure}
We also evaluated the characteristics of the action selection module. We report the percentage of actions which is selected from the knowledge based decision module (action 1) and from the reinforcement learning module (action 2) against other actions. We measured the frequency of selection of each action every 350 episodes.
As shown in Figure \ref{fig:actin selection} the action selection module at the
beginning selects equally the actions then more the action 1 and gradually give more importance to the action 2.
The results confirm our intuition, the module selects the action of the most efficient module and adapts over time the trade-off between the sources of decision to always select the best one.

\subsection{Global Model Evaluation}

\begin{figure}[tb]
\centering
\includegraphics[width=90mm, bb=0 0 697 497]{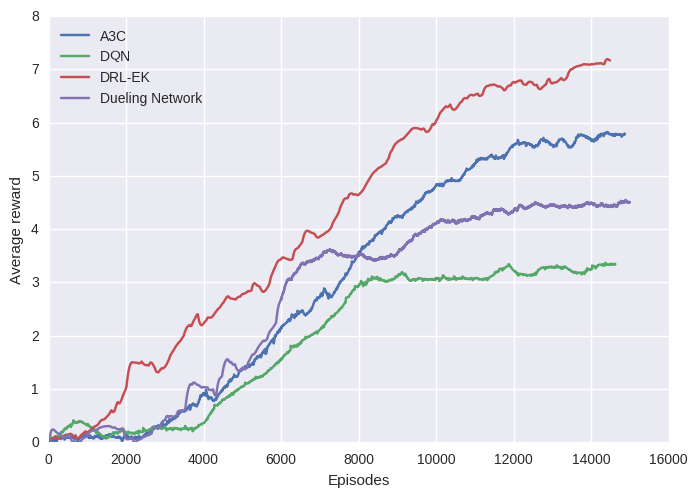}
\caption{Performance of DRL-EK comparing to DQN, Dueling Network and A3C 
\label{fig:average-reward-env1}}
\end{figure}
Finally, we report the average reward of our whole framework trained using the injection of important areas features into the reinforcement learning module and a meta-feature learning model as knowledge based decision module.
Figure \ref{fig:average-reward-env1} compares our proposed method with the best performing reinforcement
learning methods. These models learn the policy only using raw pixels.

\begin{table}
  \caption{The table compares average reward for various features injected into A3C. The reinforcement learning module was evaluated alone for 12 000 episodes.}
  \label{tab:rewardA3C}
  \begin{tabular}{ccl}
    \toprule
    Settings&Rewards\\
    \midrule
    A3C & 5.6\\
    A3C + \textit{presence of objects features} & 5.8\\
    A3C + \textit{important area features} & 6.1\\
  \bottomrule
\end{tabular}
\end{table}

DRL-EK boosts A3C by injecting important area features. To select these features, we compared \textit{A3C+presence of objects features} and \textit{A3C+important area features} (Table \ref{tab:rewardA3C}). In both cases, the results show that adding a new input to the reinforcement learning module improves the quality of the policy.

As can be seen, DQN gives the worst results with an average reward of 3.2, $\approx40\%$
less than A3C after converging.
After 12 000 episodes, the average reward of the dueling network architecture trained with a double deep Q-learning is around 4.4 while A3C is able to achieve an average reward of 5.6.
Surprisingly the meta-feature learning model trained alone (Figure \ref{fig:meta-feat_learning}) achieves higher performance than learning only from the image with a dueling network or a DQN model.

Asynchronous advantage actor-critic tends to learn faster than any other reinforcement learning based models. We believe this is due to the 3 parallel workers of A3C which offer a nonlinear significant speedup.

These results show that our architecture, DRL-EK, outperforms the baselines. Its average reward is around 15\% better than A3C after 14 000 episodes. Moreover, the performance at the beginning of the training and the learned policy of DRL-EK is significantly better than all other models.
One thing to note is that the action selection module tends to select an action different
from action 1 and 2 (Figure \ref{fig:actin selection}).
The continuous increase of the average reward of DRL-EK and this
observation indicates that the action selection module is partially able to learn to correct the errors.

The experiments demonstrate the importance of each module of our system.
With an average time for one step of 0.43 seconds on a Nvidia Titan-X (Pascal) GPU, DRL-EK can be trained in real time.

%comparaison avec A3C et A3C*
%parler presence objets

% I will add this later
%`\#\#\#\#Model presence of objects
%The human effort is limited as it only requires to generate a dataset %for the object recognition and set the object recognition threshold. %Unfortunately, it only slightly improves the performances. The %average reward is around 10\% better than without external knowledge.

\section{Conclusion}

We proposed a new architecture to combine reinforcement learning
with external knowledge. We demonstrated its ability to solve complex
tasks in 3D partially observable environments with image as input. Our central thesis is enhancing the image by generating high-level features of the environment. 
Further benefits stem from efficiently combining two sources of decision.
Moreover, our approach can be easily adapted to solve new tasks with a very limited
amount of human work.
We have demonstrated the efficacy of our architecture to decrease the training time and to learn a better and more efficient policy.

In the future, a promising research area is building an agent that incorporates human feedback. Another
challenge is how to integrate complex and structured external knowledge
such as ontologies or textual data into our model. Finally, we are interested in extending our experiments to new environments such as VizDoom \cite{kempka2016vizdoom}.

\bibliographystyle{ACM-Reference-Format}
\bibliography{sample-bibliography}

\end{document}